\documentclass[a4paper,fleqn]{cas-dc}

\usepackage{lineno,hyperref}
\modulolinenumbers[5]

\usepackage{graphics}           
\usepackage{times}              
\usepackage{amsmath}            
\usepackage{amssymb}            
\usepackage{graphicx}
\usepackage{algorithm}
\usepackage[noend]{algpseudocode}
\usepackage{booktabs}
\usepackage{color}
\definecolor{instructioncolor}{rgb}{.5,.5,.5}

\usepackage[font=small]{caption}

\def\secref#1{Sec.~\ref{#1}}
\def\figref#1{Fig.~\ref{#1}}
\def\tabref#1{Tab.~\ref{#1}}
\def\eqref#1{Eq.~(\ref{#1})}

\makeatletter
\usepackage{xspace}
\DeclareRobustCommand\onedot{\futurelet\@let@token\@onedot}
\def\@onedot{\ifx\@let@token.\else.\null\fi\xspace}
 
\def\ie{i.e\onedot}

\def\etal{{et al}\onedot}
\makeatother

\def\etalcite#1{\etal~\cite{#1}}

\usepackage{array}
\newcolumntype{L}[1]{>{\raggedright\let\newline\\\arraybackslash\hspace{0pt}}m{#1}}
\newcolumntype{C}[1]{>{\centering\let\newline\\\arraybackslash\hspace{0pt}}m{#1}}
\newcolumntype{R}[1]{>{\raggedleft\let\newline\\\arraybackslash\hspace{0pt}}m{#1}}

\renewcommand{\v}[1]{{\b #1}}

\newcommand{\q}[1]{{{\bf #1}}}

\newcommand{\cs}[2]{\; ^{#1}\!\,{\mbox{#2}}}

\usepackage{color}
\usepackage{multicol}
\usepackage{siunitx}
\usepackage{subcaption}
\usepackage{graphics}    %
\usepackage{times}       %
\usepackage{amsmath}     %
\usepackage{amssymb}     %
\usepackage{graphicx}
\usepackage{xcolor}
\usepackage[algo2e, linesnumbered,ruled,noline]{algorithm2e} 
\usepackage{multirow}

\renewcommand{\v}[1]{{\mathbf{#1}}}

\SetKwInput{KwInput}{Input}                %
\SetKwInput{KwOutput}{Output} 

\AtBeginEnvironment{algorithm}{\SetArgSty{textrm}}

\usepackage[numbers]{natbib}

\begin{document}

\let\WriteBookmarks\relax
\def\floatpagepagefraction{1}
\def\textpagefraction{.001}

\shorttitle{Adaptive Path Planning for UAVs for Multi-Resolution Semantic Segmentation}    

\shortauthors{F. Stache, J. Westheider, F. Magistri, C. Stachniss , M. Popovi\'{c}}  

\title[mode = title]{Adaptive Path Planning for UAVs\\ for Multi-Resolution Semantic Segmentation}

\author[]{Felix~Stache$^*$~\quad~Jonas~Westheider$^*$~\quad~Federico~Magistri~\quad~Cyrill~Stachniss~\quad~Marija~Popovi\'{c}$^+$}
\nonumnote{$^*$Authors with equal contribution.}

\nonumnote{$^+$Corresponding:  Marija Popovi\'{c}, Niebuhrstra{\ss}e 1A, 53113 Bonn, Germany. mpopovic@uni-bonn.de.}

\nonumnote{All authors are with the University of Bonn, Germany. This work has partially been funded by the Deutsche Forschungsgemeinschaft (DFG, German Research Foundation) under Germany's Excellence Strategy, EXC-2070 -- 390732324 -- PhenoRob and by the Federal Ministry of Food and Agriculture~(BMEL) based on a decision of the Parliament of the Federal Republic of Germany via the Federal Office for Agriculture and Food~(BLE) under the innovation support programme under funding no~28DK108B20~(RegisTer).}

\begin{abstract}
Efficient data collection methods play a major role in helping us better understand the Earth and its ecosystems. In many applications, the usage of unmanned aerial vehicles (UAVs) for monitoring and remote sensing is rapidly gaining momentum due to their high mobility, low cost, and flexible deployment. A key challenge is planning missions to maximize the value of acquired data in large environments given flight time limitations. This is, for example, relevant for monitoring agricultural fields.
This paper addresses the problem of adaptive path planning for accurate semantic segmentation of using UAVs. We propose an online planning algorithm which adapts the UAV paths to obtain high-resolution semantic segmentations necessary in areas with fine details as they are detected in incoming images. This enables us to perform close inspections at low altitudes only where required, without wasting energy on exhaustive mapping at maximum image resolution. A key feature of our approach is a new accuracy model for deep learning-based architectures that captures the relationship between UAV altitude and semantic segmentation accuracy. We evaluate our approach on different domains using real-world data, proving the efficacy and generability of our solution.\end{abstract} 

\begin{keywords}
\texttt{unmanned aerial vehicles \sep semantic segmentation \sep planning \sep terrain monitoring}
\end{keywords}

\maketitle

\section{Introduction}
\label{sec:intro}
Remote sensing and monitoring methods provide abundant data for ecology and environment research in a broad range of applications.
However, the applicability of conventional remote sensing methodologies %
in large-scale environments is limited when both fast and high-quality data collection is required.
Unmanned aerial vehicles (UAVs) are experiencing a rapid uptake in a variety of aerial monitoring applications, including search and rescue~\cite{Meera19}, wildlife conservation~\cite{Manfreda2018,Duporge2021,Ghods2021}, industrial inspection~\cite{Carrio2017,Shakhatreh2019}, and precision agriculture~\cite{popovic2017icra,popovic2017iros,Vivaldini2019,Ocer2020}. Compared to traditional data acquisition methods, such as manual or static sampling procedures \cite{Manfreda2018}, they offer a more flexible and easier to execute way to monitor areas at high spatial and temporal resolutions \cite{Manfreda2018,Shakhatreh2019}. In recent years, the advent of deep learning (DL) has unlocked their potential for image-based remote sensing, enabling flexible, low-cost data collection and processing~\cite{Carrio2017}. However, a key challenge is planning paths to efficiently gather the most useful data in large environments, while accounting for the constraints of physical platforms, e.g. on fuel/energy, as well as the characteristics of the on-board sensor and DL model used for data processing.

\begin{figure}[t]
 \centering
 \includegraphics[width=0.48\linewidth]{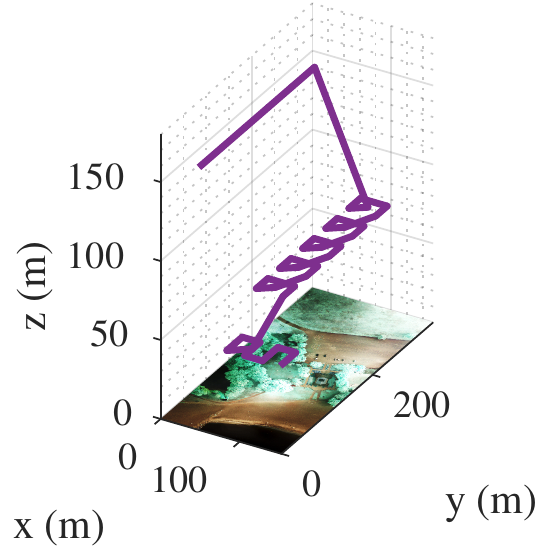}
 \includegraphics[width=0.48\linewidth]{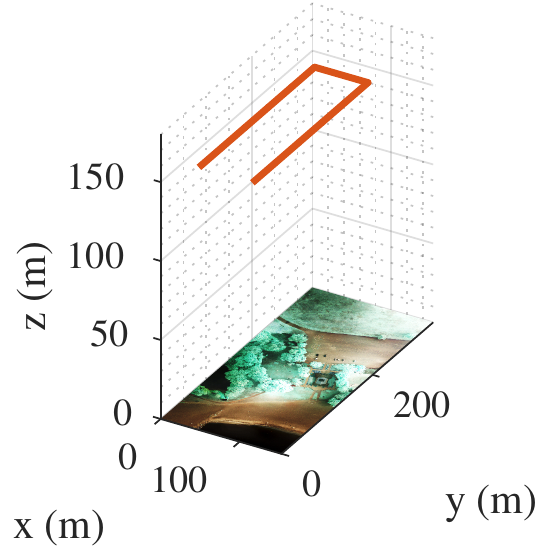}
 \vspace{1mm}

 \includegraphics[width=0.95\linewidth]{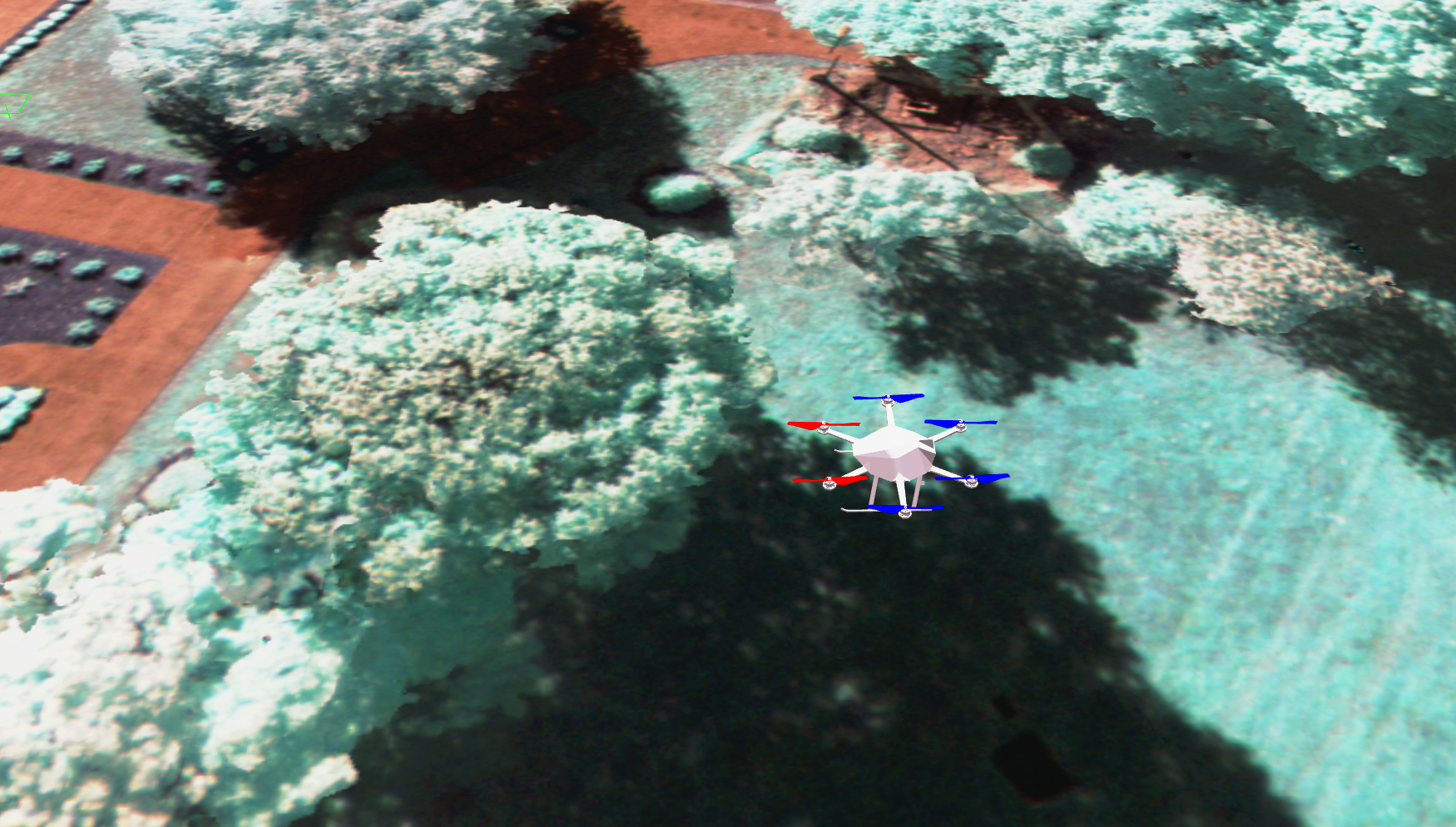}
 \caption{A comparison of our proposed adaptive path planning strategy (top-left) against lawnmower coverage planning (top-right) for UAV-based field segmentation, evaluated using real-world data from the RIT-18 dataset~\cite{kemker2018isprs} (bottom). By allowing the paths to change online, our approach enables selecting high-resolution (low-altitude) imagery in areas with more semantic detail, enabling higher-accuracy, fine-grained segmentation in these regions.}
 \label{fig:motivation}
\end{figure}

This paper examines the problem of DL-based semantic segmentation using UAVs. Specifically, we investigate how semantic information can be exploited for intelligent path planning during a mission, i.e., online. Our problem setup considers a UAV flying above a 2D terrain and taking images of it using a downwards-facing camera, as depicted in \figref{fig:motivation}. The goal is to adaptively select the next sensing locations for the UAV above the terrain to maximize the classification accuracy of objects or areas of interest seen in images, e.g. animals on grassland or crops on a field. This enables us to perform targeted high-resolution classification only where necessary and thus maximize the value of data gathered during a mission.

Aerial data acquisition campaigns rely often on coverage-based planning to generate UAV paths at a fixed flight altitude~\cite{Cabreira2019,Galceran2013}. Although they are easily implemented, the main drawback of such methods is that they assume an even distribution of features in the target environment; mapping the entire area at a constant image spatial resolution governed by the altitude. Recent work has explored \textit{informative planning} for terrain mapping, whereby the aim is to optimize an information-theoretic mapping objective subject to platform constraints. By modifying plans online, such strategies enable adapting the flight path according to the mission aim to maximize the value of collected data.
This can be used for improving the geometric estimation~\cite{palazzolo2018drones} or for semantic estimation tasks~\cite{stache2021ecmr, carbone2021dars}.
Several previous studies in UAV-based informative planning either consider planning at a fixed altitude~\cite{vivaldini2016icra,Sa2018b}, i.e. on a 2D plane above the terrain, or apply simple heuristic predictive sensor models~\cite{popovic2017icra,popovic2017iros,Meera19,Dang2018}, which limit the quality of future plans. Two open challenges are therefore: (1) reliably characterizing how the accuracy of segmented images varies with the altitude and relative scales of the objects in registered images and (2) designing strategies to incorporate such models into the planning pipeline for improved targeted data collection efficiency.  

The contribution of this paper is a new adaptive planning algorithm that directly tackles the altitude dependency of the DL semantic segmentation model using UAV-based imagery. First, our approach leverages prior labeled terrain data to empirically determine how classification accuracy varies with altitude; we train a deep neural network with images obtained at different altitudes that we use to initialize our planning strategy. 
Based on this analysis, we develop a \textit{decision function} using Gaussian Process (GP) regression that governs the decision-making strategy. This function is first initialized on a training scenario and then updated online on a spatially separate testing scenario during a mission as new images are received. 

For replanning, the UAV path is chosen according to the decision function and segmented images to obtain higher classification accuracy in more semantically detailed or interesting areas. This allows us to gather more accurate data in targeted areas without relying on a heuristic sensor model for informative planning.

This article corresponds to an extension of the authors'~preliminary conference work~\citep{stache2021ecmr}. We consolidate our previous contribution in adaptive planning with additional explanations and experimental results. A major difference is that the journal version formulates our methods in a general way, rendering them directly applicable to \textit{any} UAV-based semantic segmentation scenario, e.g. search and rescue~\cite{Meera19}, urban scene analysis, wetland assessment, in addition to precision agriculture and lake monitoring, which are studied as motivating applications in our experiments.

\section{Related Work}
\label{sec:related}

\begin{figure*}[t]
  \centering
  \includegraphics[width=0.9\linewidth]{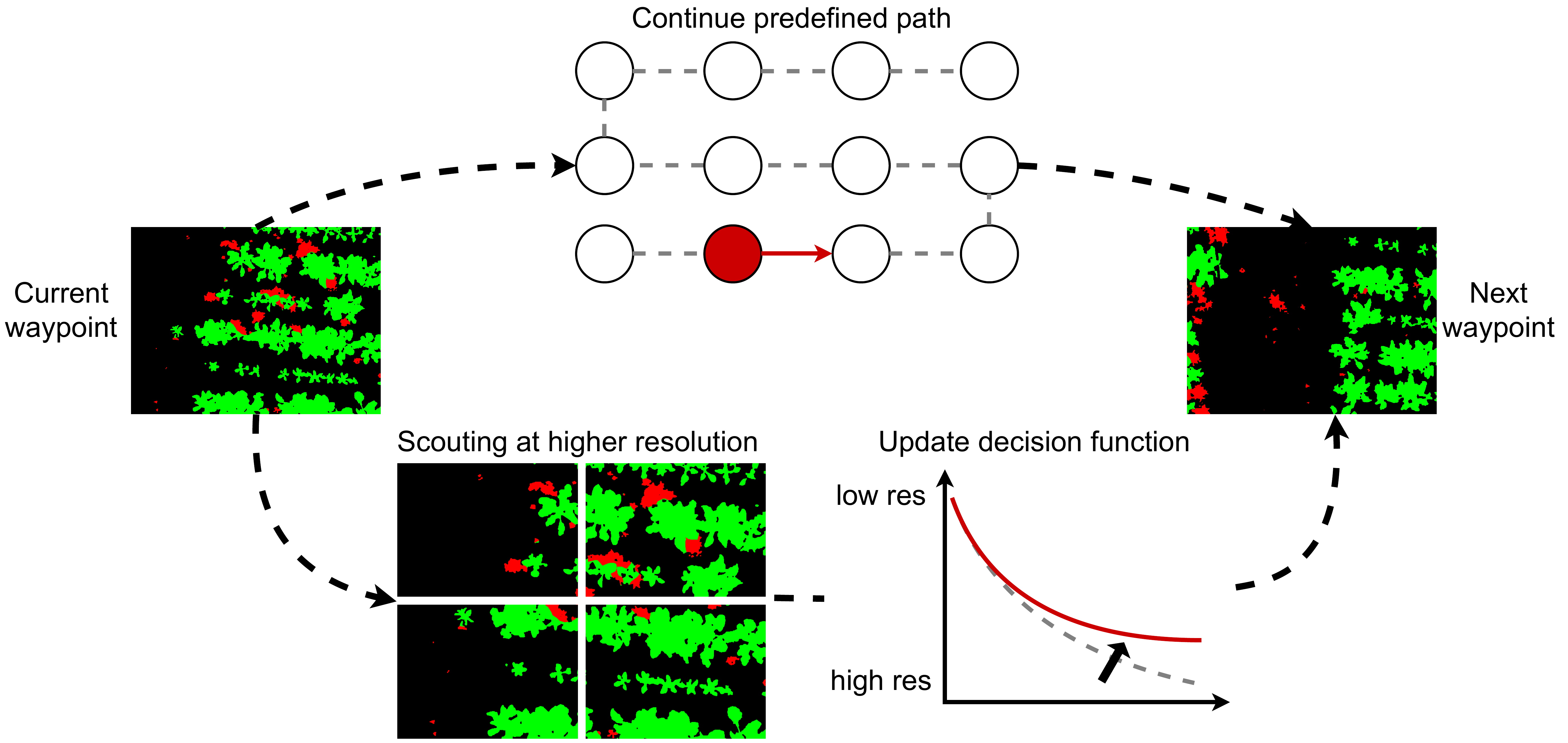}
  \caption{Overview of our adaptive planning approach. Each time one area of the field is segmented, we decide whether the UAV should follow its predefined path (`Next waypoint') or scout the same region at a lower altitude to obtain higher-resolution images here (`Scouting at higher resolution'). In the second case, we update the decision-making strategy (`Update decision function') by comparing the segmentation results of the re-observed regions at different altitudes.}
  \label{fig:workflow}
\end{figure*}

There is an emerging body of literature addressing online mission planning for UAV-based remote sensing. In this section, we briefly overview recent work related to semantic segmentation of aerial images and adaptive path planning approaches for efficient data acquisition using resource-constrained platforms.

\subsection{Semantic Segmentation Using Aerial Imagery}
The goal of semantic segmentation is to assign a predetermined class label to each pixel of an image. State-of-the-art approaches are predominantly based on fully convolutional neural networks (CNNs) due to their rich feature representation and end-to-end training capabilities, which generally allow for superior performance compared against handcrafted vision pipelines~\cite{Carrio2017}. In remote sensing, CNNs have been successfully applied to aerial image datasets in various scenarios, e.g. for crop/weed segmentation in precision agriculture~\cite{Sa2018,Sa2018b,Ocer2020}, tracking and infrastructure inspection~\cite{Nguyen2019}, urban scene analysis~\cite{Lyu2020,Avola2021}, wildlife detection~\cite{Duporge2021}, among others.

In the past few years, technological advancements have enabled computationally efficient segmentation on board small UAVs with constraints on size, weight, and power. Nguyen~\etalcite{Nguyen2019} introduced MAVNet, a light-weight network designed for real-time aerial surveillance and inspection. Sa~\etalcite{Sa2018} and Deng~\etalcite{Deng2020} proposed CNN architectures to segment vegetation for smart farming using similar platforms. Recently, Bultmann~\etalcite{Bultmann2021} designed a UAV system for real-time semantic fusion using multiple sensor modalities. Our work examines a problem setup similar to these studies: we adopt the light-weight ERFNet architecture, introduced by Romera~\etalcite{romera17} for semantic segmentation of terrain based on aerial images obtained from a downwards-facing camera.

A fundamental difference with respect to the works above is that, instead of flying predetermined paths for data collection, our focus is on adaptive planning. We propose to exploit accurate real-time segmentation capabilities to modify the flight plan online to achieve targeted data acquisition. Specifically, our goal is to localize areas of interest and finer detail (e.g. high vegetation cover in a field or victims in a disaster site) online and steer the robot for adaptive, high-accuracy mapping in these regions.

\subsection{Multi-Resolution Monitoring}
An important trade-off in aerial imaging arises from the fact that the same point on a field can be observed from different altitudes. As a result, image spatial resolution degrades with increasing ground area coverage, i.e. the closer a UAV flies to the ground plane, the greater the level of image detail, but the smaller the observed area and hence the higher the flight time required to completely cover a field of fixed size. Pe{\~{n}}a~\etalcite{Pena2015} established that there are optimal altitudes for monitoring plants based on their size. Duporge~\etalcite{Duporge2021} presented similar findings in wildlife monitoring applications. These studies motivate our approach for adaptively modifying the flight altitude during a mission based on the image content.

Relatively limited research has addressed the altitude-resolution trade-off in the contexts of semantic segmentation and robotic motion planning. Several works \cite{Avola2021,Ocer2020} investigated network architectures with variable-size kernels that are robust to flight altitude changes. However, such methods have not yet been studied in adaptive decision-making scenarios where the physical robot constraints during data collection are taken into account.

Various methods have been proposed to address planning with multi-resolution sensors. For 3D mapping with image-based semantic information, Dang~\etalcite{Dang2018} employed an interesting method for weighting distance measurements according to their resolution. This sensor model is used to guide exploration planning in unknown environments based on the current map state. Sadat~\etalcite{Sadat15} proposed an adaptive coverage-based strategy that assumes sensor accuracy increases with altitude. Other studies~\cite{Vivaldini2019,Sa2018b} only considered fixed-altitude mission planning in terrain monitoring problems; thereby neglecting the altitude dependency of the camera. We follow previous approaches that empirically assess the effects of multi-resolution observations for trained models~\cite{Meera19,Popovic2020AURO,Qingqing2020,Ghods2021}. In contrast to these works, which derive the sensor model for planning offline, i.e. before a mission, our contribution is a decision function that supports \textit{online} updates based on incoming images for more reliable predictive planning performance. 

\subsection{Adaptive Path Planning}
Adaptive algorithms for active sensing allow an agent to replan online as measurements are collected during a mission to focus on application-specific interests. Several works have successfully incorporated adaptivity requirements within informative path planning problems. Here, the objective is to minimize uncertainty in target areas as quickly as possible, e.g. for exploration~\cite{stachniss2005rss,Dang2018}, underwater surface inspection~\cite{Hollinger2013}, target search~\cite{Meera19,Sadat15,Singh2009,Ghods2021}, and environmental sensing~\cite{Singh2009,Popovic2020AURO}. These problem setups differ from ours in several ways. First, they consider a probabilistic map to represent the entire environment, using a sensor model to update the map with new uncertain measurements. In contrast, our approach directly exploits the accuracy in semantic segmentation to drive the next actions in adaptive planning. This circumvents the computational expenses of storing and updating a global map. Second, they consider a predefined, i.e. non-adaptive, sensor model, whereas ours is adapted online using a GP according to the behavior of the semantic segmentation model. The usage of GPs for path planning has already been exploited by Nardi~\etalcite{nardi2019icra-airn}; in contast to their method, we additionally use online segmentation to adapt the GP online.

Very few works have considered planning based on semantic information. O{\ss}wald~\etalcite{osswald2016ral} proposed to explore a scene exploiting background information given by a user. Bartolomei~\etalcite{Bartolomei20} introduced a perception-aware planner for UAV pose tracking. Although, like us, they exploit semantics to guide next UAV actions, their goal is to triangulate high-quality sparse landmarks whereas we aim to obtain accurate pixel-wise semantic segmentation in dense images. Dang~\etalcite{Dang2018} and Meera~\etalcite{Meera19} studied informative planning for active target search using object detection networks with distance-based uncertainties. Ghods~\etalcite{Ghods2021} explored a similar problem setup with multiple search agents. Most similar to our approach is that of Popovi\'{c}~\etalcite{Popovic2020AURO}, which adaptively plans the 3D path of a UAV for terrain monitoring based on an empirical performance analysis of a SegNet-based architecture at different altitudes~\cite{Sa2018}. A key difference is that our decision function, representing the network accuracy, is not static. Instead, we allow it to change online and thus adapt to new unseen environments. Moreover, for path planning, we present a general approach applicable for problems with different numbers of semantic labels.

\section{Our Adaptive Path Planning}
\label{sec:main}

Our problem setup considers a UAV surveying a flat field of known size using a downwards-facing camera and subject to flight time constraints. The goal is to maximize the accuracy in the semantic segmentation of RGB images taken by the camera. We propose a data-driven approach that uses information from incoming images to adapt an initial predefined UAV flight path online. The main idea behind our approach is to guide the UAV to take high-resolution images for fine-grained segmentation at lower altitudes (higher spatial resolutions) only in areas where high semantic detail is desired.

\figref{fig:workflow} shows an overview of our planning strategy. %
We first divide the target field into non-overlapping regions and, for each, associate a waypoint in the 3D space above the field from which the camera footprint of the UAV camera covers the entire area. From these waypoints, we then define a lawnmower coverage path that we use to bootstrap the adaptive strategy. Our planning strategy consists of two main steps. First, at each waypoint along the lawnmower path, we use a deep CNN to assign a semantic label to each pixel in the region observed in the image. %
Second, based on the segmentation output, we decide whether the current region contains enough semantic value for more detailed re-observation at a higher image resolution, \ie, lower UAV altitude; otherwise, the UAV continues its pre-determined coverage path. The replanning procedure is repeated for each waypoint on the original lawnmower path.

A key aspect of our approach is a new data-driven decision function which enables the UAV to select a new altitude for higher-resolution images only if they are needed.
The decision function captures the relative pixel-wise ratio of semantic labels of interest in an image, allowing us to judge whether an area contains high semantic value and thus needs closer inspection. This function is updated adaptively during the mission by comparing the segmentation results of the current region at the different altitudes. In this way, we can precisely capture the relationship between image resolution (altitude) and segmentation accuracy when planning new paths.

In the following sub-sections, we first describe the CNN for semantic segmentation (\secref{subsec:segmentation}) before detailing our path planning strategy, which consists of offline planning (\secref{subsubsec:init} and \secref{subsubsec:decision}) and online path adaptation (\secref{subsubsec:online}).

\subsection{Semantic Segmentation}
\label{subsec:segmentation}
In this work, we consider the semantic segmentation of RGB images not only as of the final mission goal but also as the tool within our planning algorithm used to define adaptive paths for re-observing given regions of the field.
Each time the UAV reaches a waypoint, we perform pixel-wise semantic segmentation in the current view to assign a class label to each pixel from the set $ \mathcal{C} = \{ l_1, l_2,\ldots, l_C \} $, where $C$ is the number of classes.
Specitically, our proposed approach leverages the ERFNet~\cite{romera17} architecture provided by the Bonnetal framework~\cite{milioto2019ieee} that allows for real-time inference. We train this CNN on RGB images collected at different altitudes to allow it to generalize across possible altitudes without the need for retraining. 
If the same region is observed by the camera from different altitudes, we preserve the results obtained with the highest resolution, assuming that higher-resolution images yield greater segmentation accuracy.

\begin{figure}[t]
  \centering
   \includegraphics[width=0.99\linewidth]{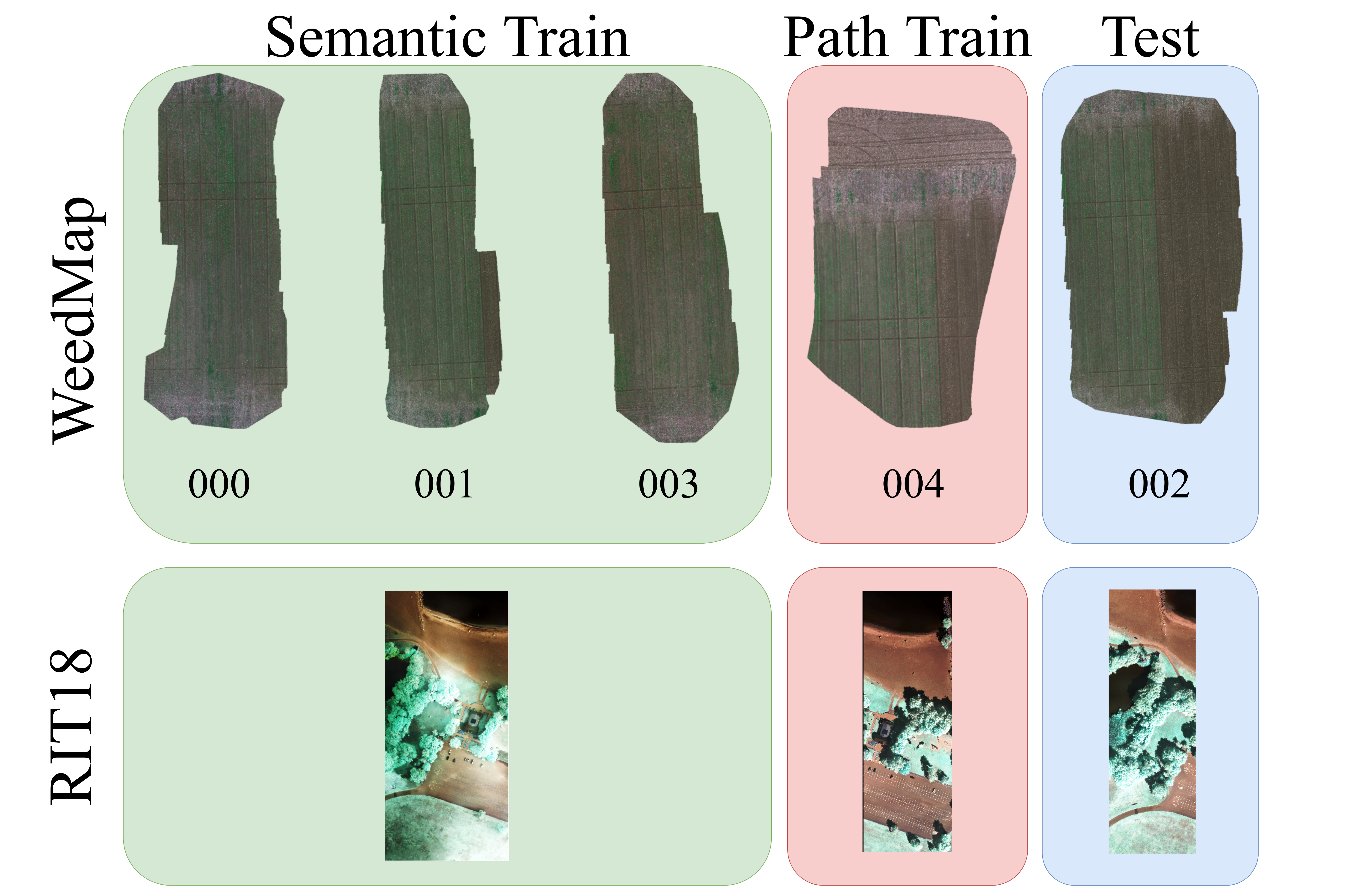}
  \caption{Our experimental setup using the WeedMap \cite{Sa2018b} and RIT-18 \cite{kemker2018isprs} datasets. Green, red, and blue indicate the fields used to train a CNN for semantic segmentation, initialize the planning strategy, and for evaluation. For an extensive evaluation of our approach, we swap the roles of the fields so that we test our algorithm on each field once.}
  \label{fig:dataset_split}  
\end{figure}

\subsection{Path Planning}
\label{subsec:planning}

Given a trained CNN model, our path planning algorithm can be divided in three parts. First, we define a lawnmower strategy to cover the entire region of interest. Second, we initialize a decision function, based on the data obtained in previous flights on a spatially disjoint region, which serves as the starting point for our online planner. Third, while flying above the region of interest, we update the decision function as soon as new data is available.

\subsubsection{Initial Strategy}
\label{subsubsec:init}

In the first replanning step, the initial UAV flight path is calculated at a fixed altitude above the target field based on a standard zig-zag lawnmower strategy~\cite{Galceran2013}. Such a path enables covering the field efficiently assuming no prior knowledge about it is available. In \secref{subsubsec:online}, we adapt this initial path according to the non-uniform distribution of features of interest to improve semantic segmentation performance.

For a desired region of interest, we define a lawnmower path based on a series of waypoints. A waypoint is defined as a position $\v{w}_i$ in the 3D UAV workspace above the field where: (i) the UAV camera footprint does not overlap the footprints of any other waypoint; (ii) the UAV performs the semantic segmentation of its current field of view; (iii) the UAV decides to revise its path or to execute the path as previously determined; and (iv) we impose zero velocity and zero acceleration.

The initial flight path is calculated in form of fixed waypoints at the highest altitude $ \v{W}^{h_\text{max}} = \{ \v{w}_0, \v{w}_1,\ldots, \v{w}_n \} $, which is empirically set. If necessary, we modify this coarse plan by inserting further waypoints based on the new camera imagery as it arrives. At each waypoint $\v{w}_i$, the UAV decides either to follow the pre-computed coverage path, \ie moving to $\v{w}_{i+1}$, or to inspect the current region more closely at a lower altitude. In the second case, we define a second series of waypoints, $ \v{W}^{h^\prime} = \{ \v{w}_0, \v{w}_1, \ldots, \v{w}_n \} $, at the desired altitude, $h^\prime$, that will be inserted before $\v{w}_{i+1} \in \v{W}^{h_\text{max}}$ so that the resulting path, at the desired altitude, is a lawnmower strategy covering the camera footprint from $\v{w}_i \in \v{W}^{h_\text{max}}$.

\subsubsection{Decision Function Initialization}
\label{subsubsec:decision}

We develop a decision function that takes a given waypoint as input and outputs the next waypoint, either $\v{w}_{i+1} \in \v{W}^{h_\text{max}}$ or $\v{w}_0 \in \v{W}^{h^\prime}$, given the semantic segmentation result. In the second case, where there is an altitude change, our decision function also outputs the value of the desired altitude $h^\prime$.
To do this, we start by defining a subset of class labels $ \mathcal{L} = \{ l_1, l_2,\ldots, l_L \}, \mathcal{L} \subseteq \mathcal{C} $ considered as being interesting for more detailed semantic analysis. For a segmented image, we compute the number of pixels belonging to these labels as a fraction of the total number of pixels:
\begin{equation}\label{eq:vr} 
\sigma = \frac{\sum_{l \in \mathcal{L} } p_l }{P_{\text{tot}}} \,,
\end{equation}
where $P_{tot}$ is the total number of pixels in the image and $p_l$ is the total number of pixels classified as the labels in $\mathcal{L}$. The \textit{semantic ratio} $\sigma$ gives us a way to infer how valuable it is to spend time on the current region of the field. It captures the intuition that higher values of this ratio indicate more possible misclassifications among the class labels of interest. To quantify such a relationship, we let the UAV run on a separate field, where we have access to ground truth data, segmenting regions of the fields with different altitudes. Segmenting the same region of the field with different altitudes provides two pieces of information that we use to shape the decision function. On one hand, we have the difference between the altitudes from which we segment the field, $\Delta h = h_\text{max} - h^\prime$. On the other hand, we have the the difference between the semantic ratio $\sigma$ in the predicted segmentation, $\Delta \sigma = \sigma_{h_\text{max}} - \sigma_{h^\prime}$. At the same time, we can compare $\sigma$ to the accuracy of the predicted segmentation by computing the mean intersection over union (mIoU). The mIoU is defined as the average over the classes $\mathcal{L}$ of the semantic ratio between the intersection of ground truth (gt) and predicted segmentation (pred) and the union of the same quantities:

\begin{equation}\label{eq:iou}
\text{mIoU} = \frac{1}{|\mathcal{L}|} \sum_{l \in \mathcal{L}}  \frac{\text{gt}_l \cap \text{pred}_l}{\text{gt}_l \cup \text{pred}_l}\,.
\end{equation}
Again, we define the difference between mIoUs at different altitudes as $\Delta \text{mIoU} = \text{mIoU}_{h_\text{max}} - \text{mIoU}_{h^\prime}$ .

Our method thus considers two sets of observations, representing the relationships between the semantic ratio $\sigma$ and UAV altitude $h$ (called $\mathcal{O}$) and between the ratio $\sigma$ and mIoU (called $\mathcal{I}$) as follows:
\begin{equation}\label{eq:veg2h}
  \mathcal{O} = \begin{bmatrix}
   \Delta \sigma_0 & \Delta h_0\\ 
   \Delta \sigma_1 & \Delta h_1\\ 
   \multicolumn{2}{c}{\vdots}\\
   \Delta \sigma_n & \Delta h_n\\ 
  \end{bmatrix}, \,\,\,\,\,
  \mathcal{I} = \begin{bmatrix}
   \Delta \sigma_0 & \Delta \text{mIoU}_0\\ 
   \Delta \sigma_1 & \Delta \text{mIoU}_1\\ 
   \multicolumn{2}{c}{\vdots}\\
   \Delta \sigma_n & \Delta \text{mIoU}_n\\ 
  \end{bmatrix}.
\end{equation}

While both sets are initialized offline, we only update $\mathcal{O}$ online given that $\mathcal{I}$ requires access to ground truth data, which is clearly not available on testing fields. We fit both sets of observations using Gaussian Process (GP) regression, a nonparametric Bayesian regression approach \cite{Rasmussen2006}. A GP assumes a Gaussian process prior~$f(x)$, which is fully defined by a mean function~$m(x)$ and a covariance function~$k(x_i,\,x_j)$:

\begin{equation}\label{eq:prior}
f(x) \thicksim \mathcal{GP}(m(x),\,k(x_i,x_j))\,.
\end{equation}

To capture environmental phenomena, a common choice is to set the mean function~$m(x) = 0$ and to use the squared exponential covariance function:
\begin{eqnarray}
\label{eq:sqexp}
  k(x_i,x_j) &=& \varsigma_{f}^2 \mathrm{exp}\left(-\frac{1}{2}\frac{|x_i-x_j|^2}{\ell^2}\right) +\varsigma_{n}^2,
\end{eqnarray}
\noindent where $\mathbf{\theta} =\{\ell,\,\varsigma_{f}^2,\,\varsigma_{n}^2\}$ are the model hyperparameters and represent respectively the length scale~$\ell$, the variance of the output~$\varsigma_{f}^2$ and of the noise~$\varsigma_{n}^2$.
Typically, the hyperparameters are learned from the training data by maximizing the log marginal likelihood.
Given a set of observations~$y$ of~$f$ for the inputs~$\v{X}$ (\ie our sets $\mathcal{O}$, $\mathcal{I}$), GP regression allows for learning a predictive model of~$f$ at the query inputs~$\v{X}_{*}$ by assuming a joint Gaussian distribution over the samples. 
The predictions at~$\v{X}_{*}$ are represented by the predictive mean~$\mu_{*}$ and variance~$\sigma_{*}^2$ defined as:

\begin{eqnarray}
  \begin{aligned}
    \mu_{*} =&~\v{K}(\v{X}_{*},\v{X})\,\v{K}_{\mathrm{XX}}^{-1}\,y, \\
    \sigma_{*}^2 =&~\v{K}(\v{X}_{*},\v{X}_{*})\,-\v{K}(\v{X}_{*},\v{X})\,{\v{K}_{\mathrm{XX}}}^{-1}\,\v{K}(\v{X},\v{X}_{*}),
  \end{aligned}
  \label{eq:gp}
\end{eqnarray}

\noindent where~$\v{K}_{\mathrm{XX}} = \v{K}(\v{X},\v{X})+\varsigma_{n}^2\,\v{I}$, and $\v{K}(\cdot,\,\cdot)$ are matrices constructed using the covariance function~$k(\cdot,\cdot)$ evaluated at the training and test inputs, $\v{X}$ and~$\v{X}_{*}$. In the following, we will use the ground sampling distance (GSD) to identify the image resolution (thus the UAV altitude) from which semantic segmentation is performed. The GSD is defined as: $\text{GSD} = \frac{h \* S_w}{f \* I_w }$, where $h$ is the UAV altitude in meters, $S_w$ is the camera sensor width in centimeters, $f$ the focal length of the camera in centimeters and $I_w$ is the image width in pixels.

\subsubsection{Online Adaptation}
\label{subsubsec:online}
To adapt the UAV behavior online to fit the differences between the testing and training fields, we update the GP defined by the set $\mathcal{O}$ in the following way. In the testing field, each time the UAV decides to change altitude to a lower one, we compute a new pair $\Delta \sigma' , \Delta h'$ and re-compute the GP output as defined in \eqref{eq:prior}. This procedure is repeated for each waypoint on the original lawnmower path. 

\begin{figure*}[t]
  \centering
   \includegraphics[width=0.3\linewidth]{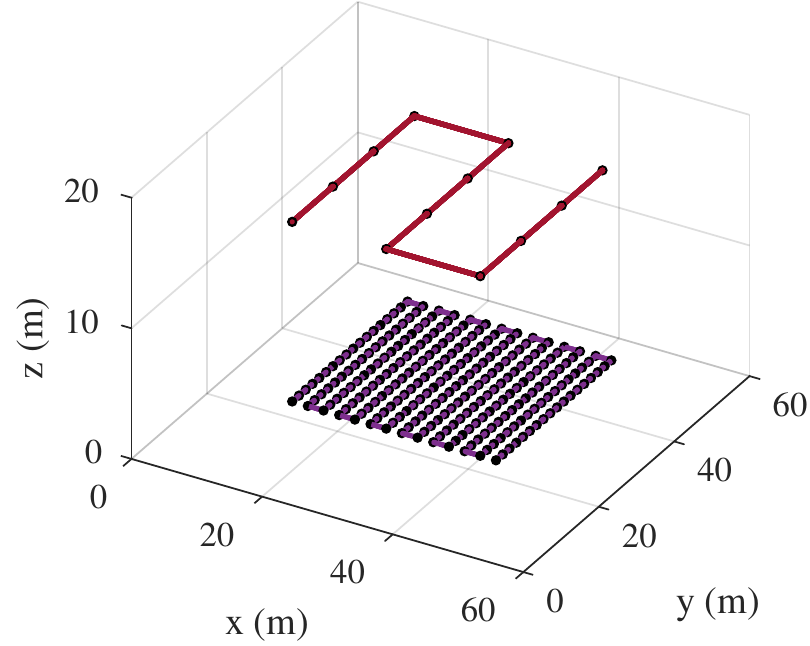}
   \includegraphics[width=0.3\linewidth]{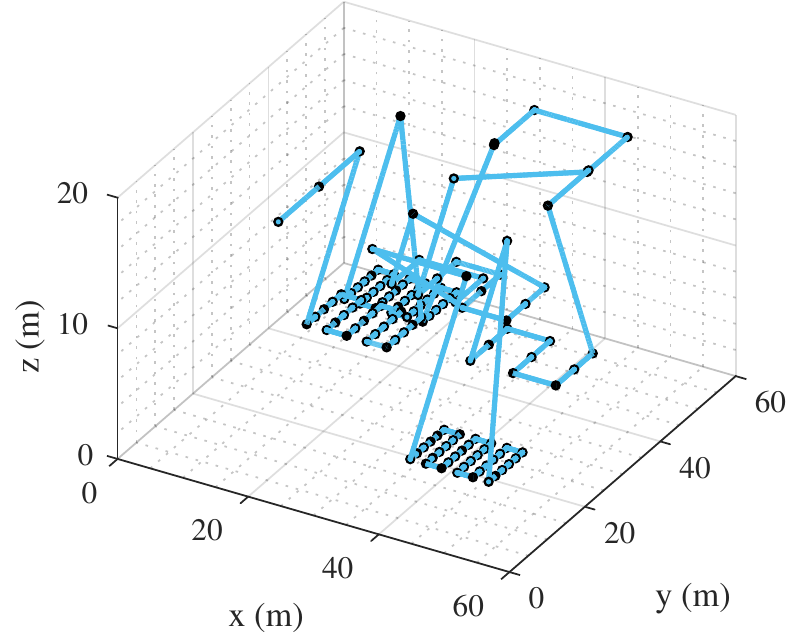}
   \includegraphics[width=0.3\linewidth]{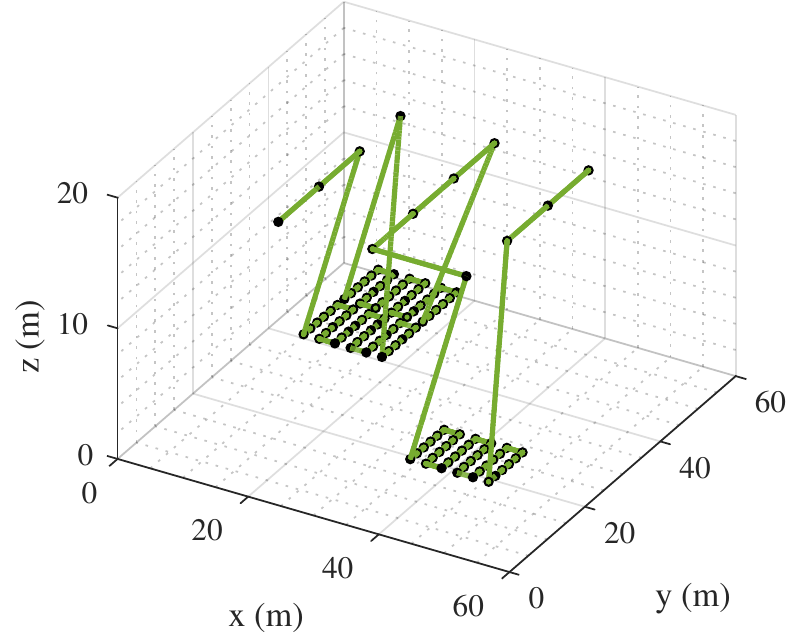}\\
   \includegraphics[width=0.3\linewidth]{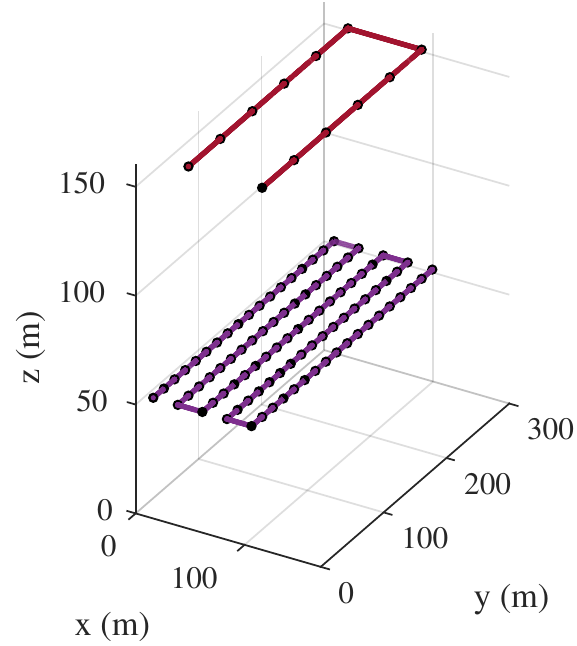}
   \includegraphics[width=0.3\linewidth]{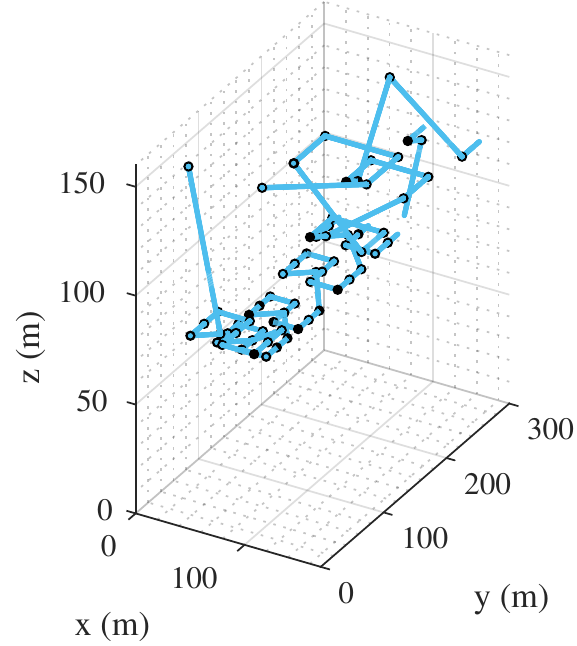}
   \includegraphics[width=0.3\linewidth]{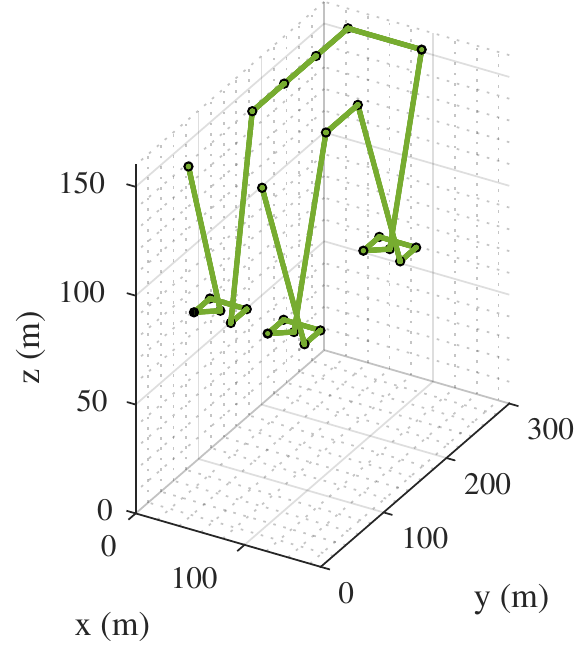}
  \caption{Visual comparison of trajectories traveled by the UAV over a field using different planning strategies. Top WeedMap, bottom RIT-18. The coverage paths (left) are restricted to fixed heights and cannot map targeted areas of interest. The linear decision function (middle) enables adaptive planning, but it is continuous with respect to altitude and leads to sudden jumps. Our adaptive approach overcomes this issue, leaving the path less often and more purposefully at selected heights for more efficient mapping. The black spheres indicate measurement points.}
  \label{fig:waypoints}
\end{figure*}

\section{Experimental Results}
\label{sec:experimental}

We validate our proposed algorithm for online adaptive path planning on the application of UAV-based semantic segmentation. The goal of our experiments is to demonstrate the benefits of using our adaptive strategy to maximize segmentation accuracy in missions while keeping a low execution time. Specifically, we show results to support two key claims: our online adaptive algorithm can (i) map high-interest regions with higher accuracy and (ii) improve segmentation accuracy while keeping a low execution time with respect to the baselines described in \secref{subsec:baselines}.

\subsection{Datasets}

Our approach is evaluated using two real-world datasets, WeedMap~\cite{Sa2018b} and RIT-18~\cite{kemker2018isprs}. We consider aerial data captured from different domains to demonstrate the general applicability of our method. WeedMap consists of 8 different fields collected with two different having different channels. It also provides pixel-wise semantic segmentation labels for each of the 8 fields; the class labels present in this dataset are \textit{soil}, \textit{crop} and \textit{weed}. In this study, we focus only on the 5 fields having RGB information. We split the 5 fields into training and testing sets (\figref{fig:dataset_split}). One of the training fields is used to initialize the decision function that shapes altitude selection in the adaptive strategy, as described in \secref{subsubsec:online}. RIT-18 consists of labelled high-resolution multi-spectral orthomosaics obtained from remote sensing imagery, in this dataset we consider the labels for \textit{asphalt}, \textit{beach}, \textit{vegetation}, \textit{water}, and \textit{building}. Note that, in order to reduce the complexity of semantic segmentation task for our experimental purposes, we define such labels by grouping together similar classes in the original dataset.

\figref{fig:dataset_split} specifies the the dataset splits studied in our experimental setup. 
For each experiment in the following sub-sections, we test our approach and the baselines described in \secref{subsec:baselines} on each field once, and then report the average values for each run.

\subsection{Baselines}
\label{subsec:baselines} 
To evaluate our proposed approach, we compare it against two main baselines.

The first one is the standard lawnmower strategy where a UAV covers the entire field at the same altitude, for this strategy we use consider five different altitudes resulting in GSD $\in \{ 1.0, 1.5, 2.0, 2.5, 3.0 \}~\text{cm}/\text{px}$. The lawnmower strategy with a fixed GSD of $3.0~\text{cm}/\text{px}$ corresponds to the initial plan for our strategy described in \secref{subsec:planning}.
The second baseline is defined by \textit{only} initializing the UAV behavior as described in \secref{subsubsec:init} and without adapting the strategy online using the decision function as new segmentations arrive. We refer to this strategy as ``Non Adaptive''. This benchmark allows us to study the benefit of adaptivity obtained by using our proposed approach (``Adaptive'').

\begin{figure}
     \centering
     \begin{subfigure}[b]{\linewidth}
      \centering 
      \includegraphics[width=\linewidth]{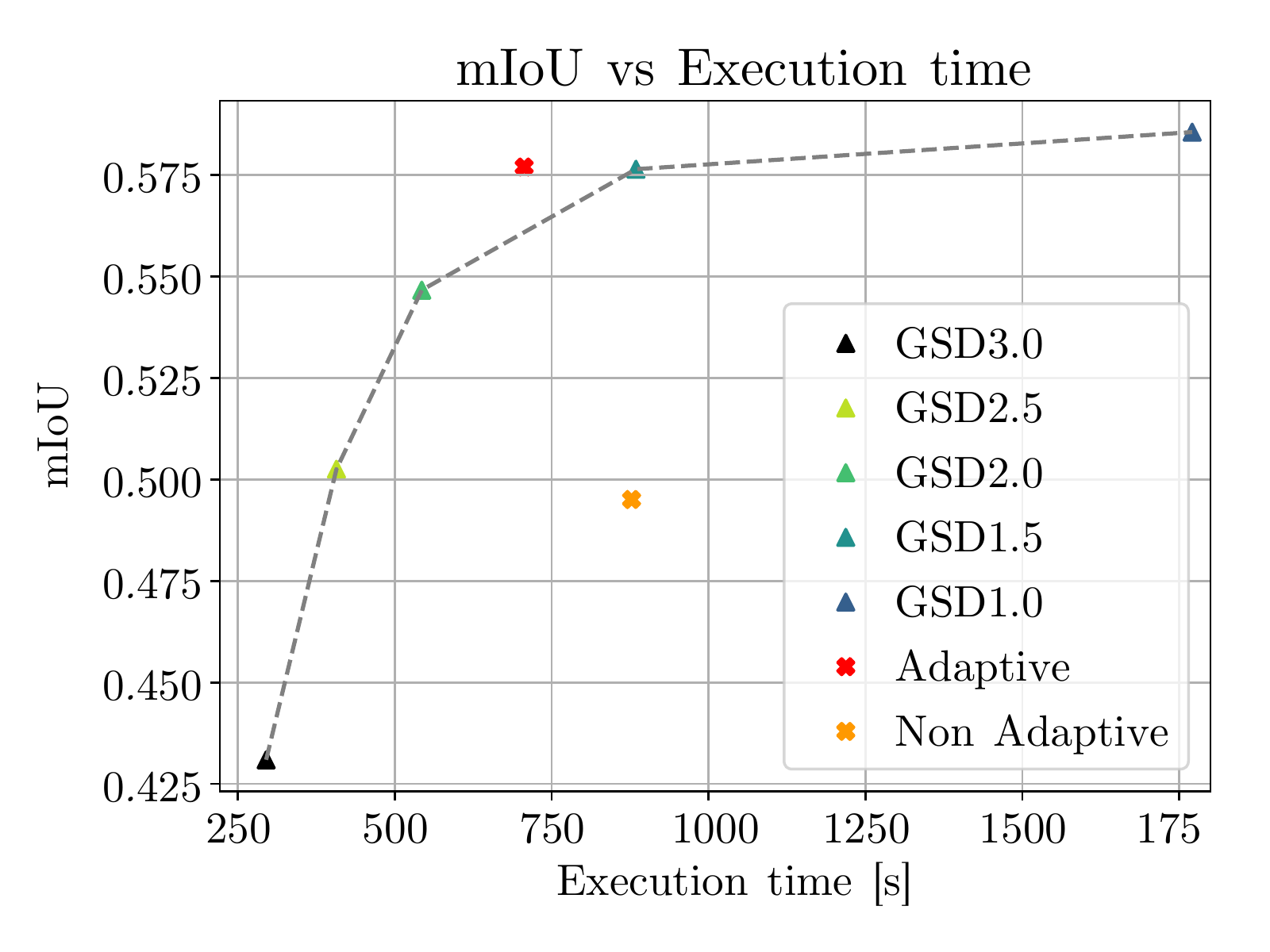}
      \caption{WeedMap.}
      \label{fig:results}
     \end{subfigure}
     \hfill
     \begin{subfigure}[b]{\linewidth}
      \centering 
      \includegraphics[width=\linewidth]{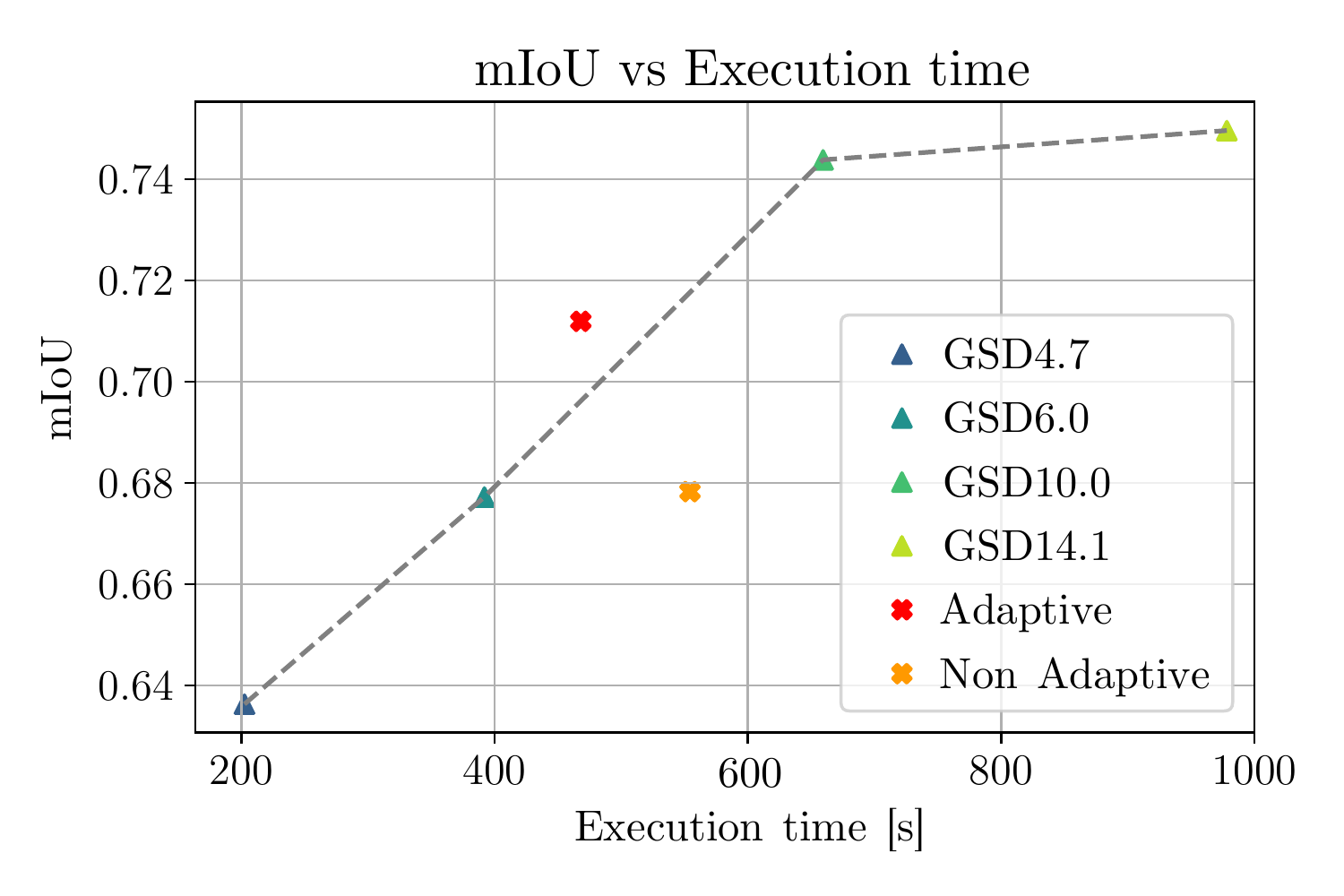}
      \caption{RIT-18.}
      \label{fig:results_rit}
     \end{subfigure}
      \caption{Averaged results for the testing fields. The red cross lies to the left of all performances with a linear decision function, indicating performance improvement.}
\end{figure}

\subsection{Metrics}
\label{subsec:metrics}

Our evaluation considers two main criteria: segmentation accuracy and mission execution time. For execution time, we compute the total time taken by the UAV to survey the whole field, including the time needed to move between waypoints, segment a new image, and plan the next path. To assess the quality of the semantic segmentation we compute the mIoU metric according to~\eqref{eq:iou}.

\begin{figure}
\centering
\begin{subfigure}{.5\linewidth}
  \centering
  \includegraphics[width=.99\linewidth]{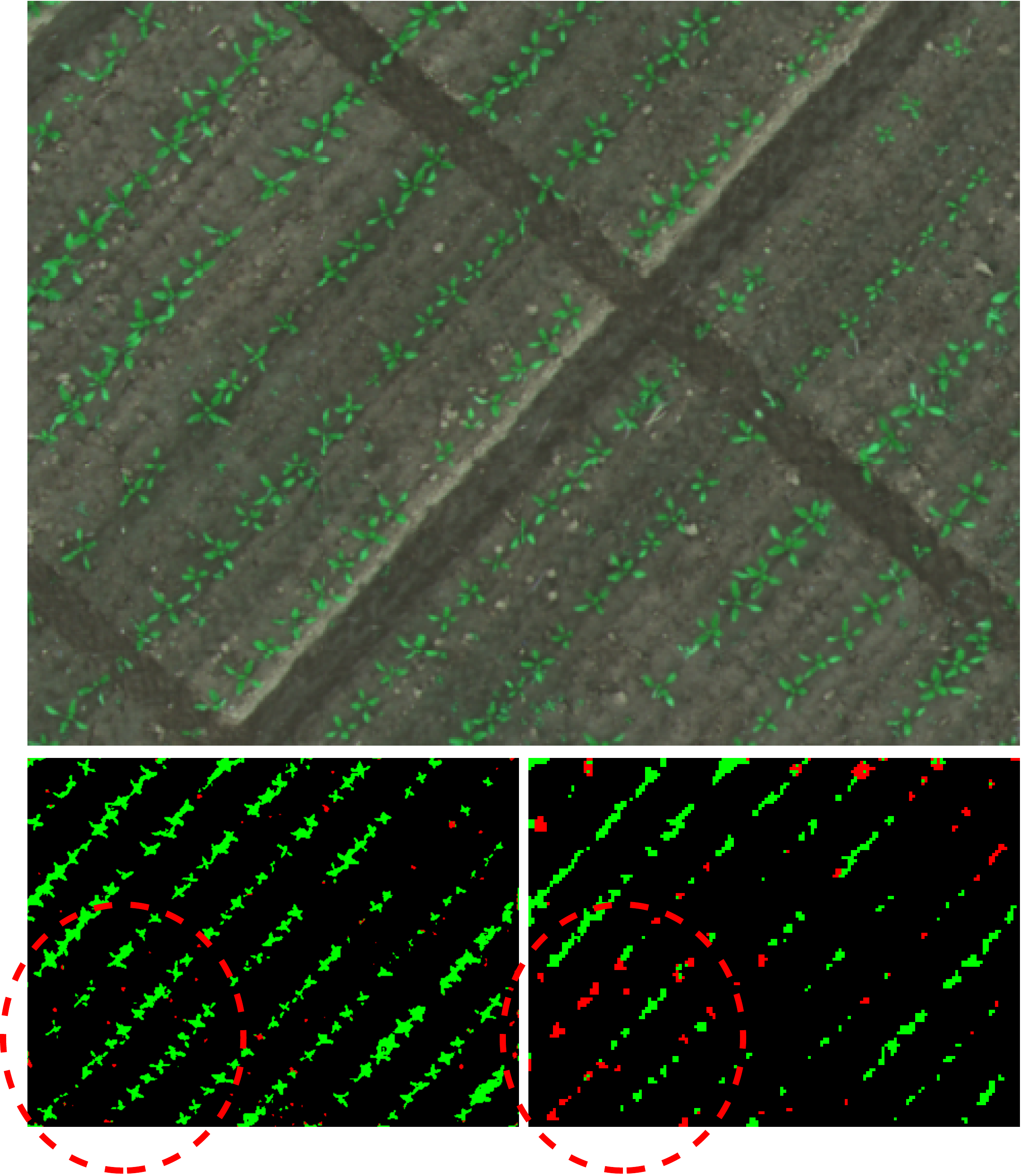}
  \caption{WeedMap.}
  \label{fig:results2}
\end{subfigure}%
\begin{subfigure}{.5\linewidth}
  \centering
  \includegraphics[width=.96\linewidth]{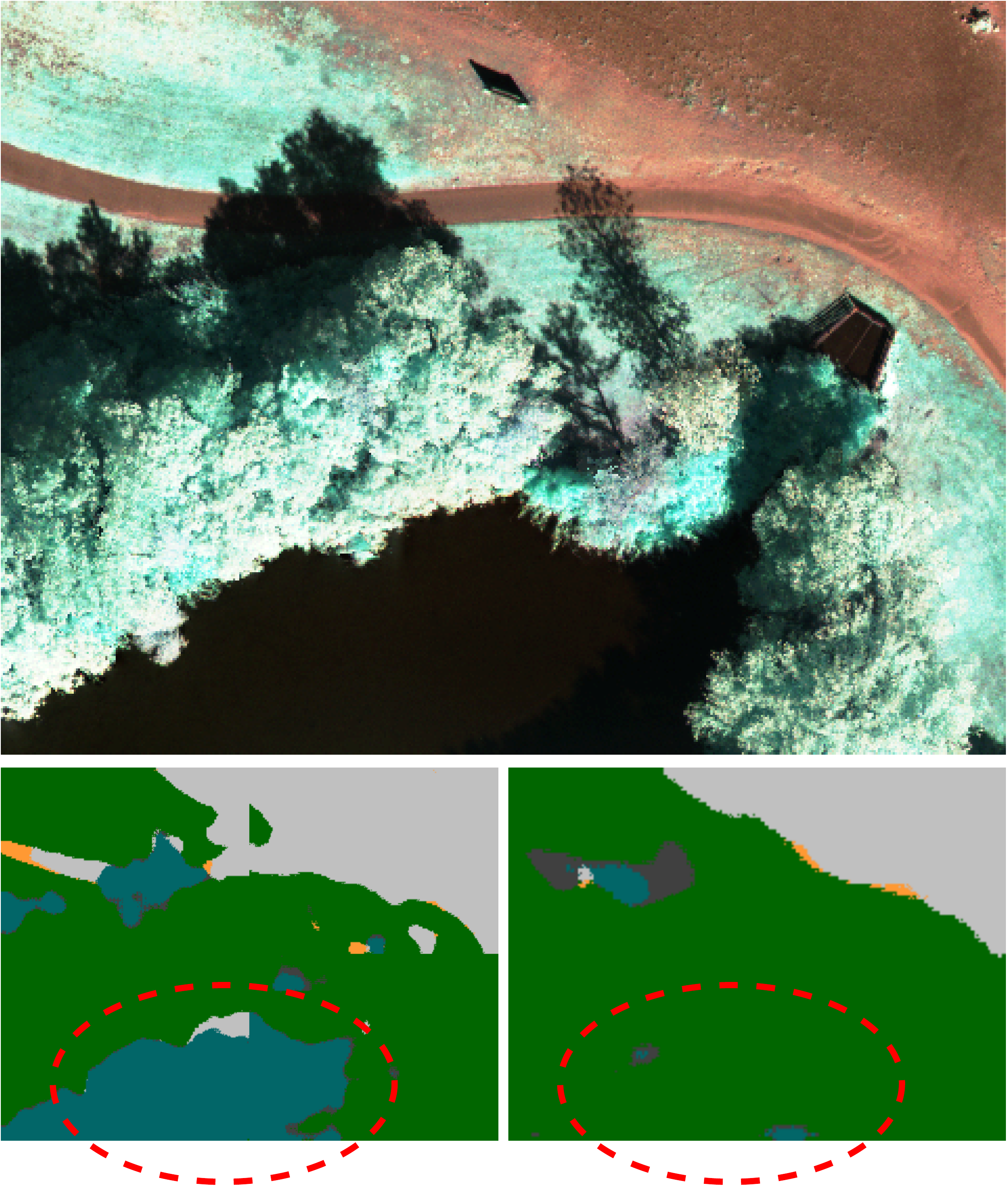}
  \caption{RIT-18.}
  \label{fig:results2_rit}
\end{subfigure}
\caption{Qualitative field segmentation results using the proposed adaptive strategy using our decision function (bottom left) and lawnmower strategy (bottom right) for path planning. The circled details demonstrate that our adaptive planning approach enables targeted high-resolution segmentation to capture finer details at higher accuracy.}
\label{fig:qualitative}
\end{figure}

\subsection{Field Segmentation Accuracy vs. Execution Time}
\label{sec:semseg_acc}

\begin{figure}
     \centering
      \begin{subfigure}[b]{\linewidth}
        \centering
         \includegraphics[width=0.9\linewidth]{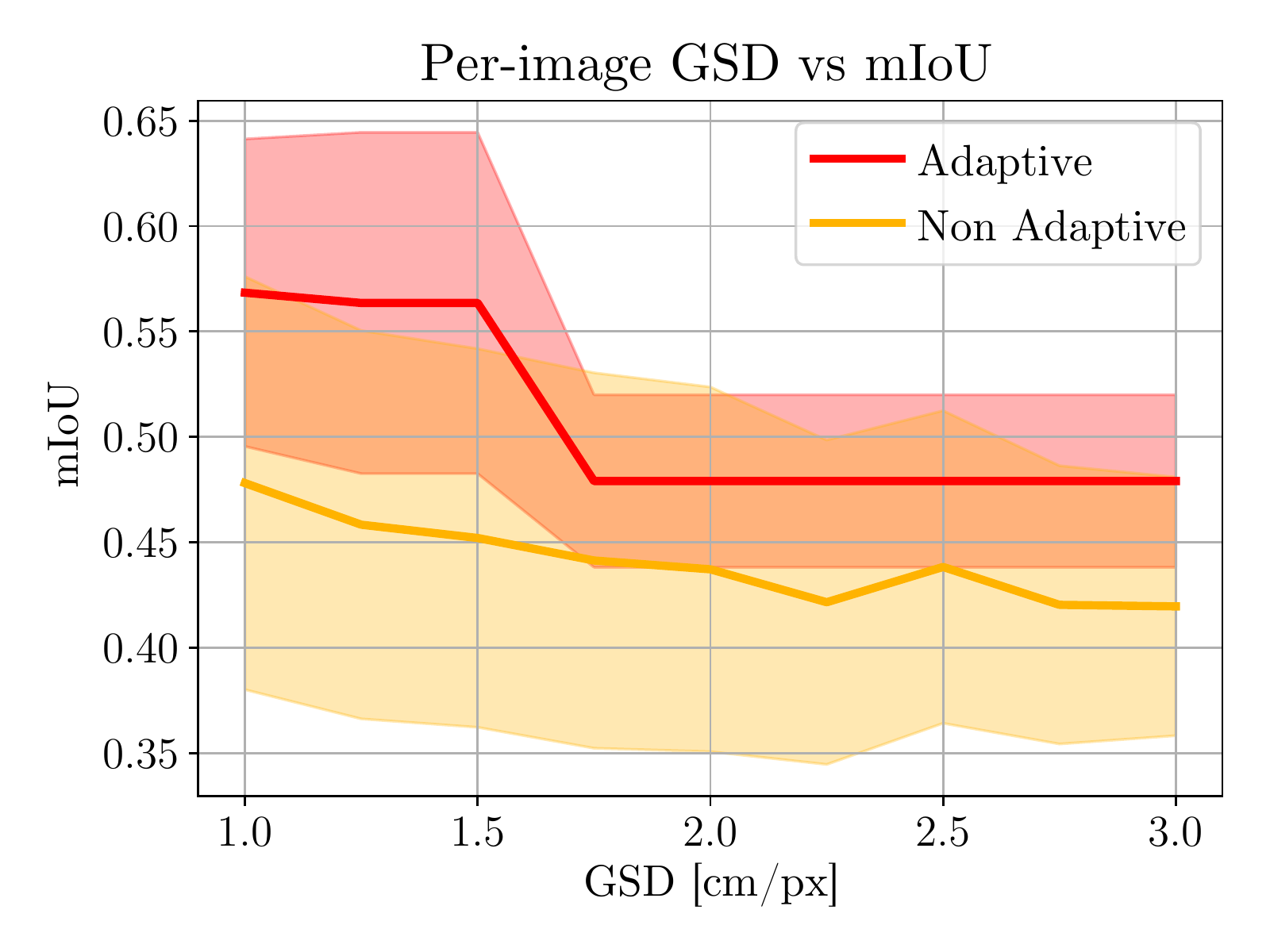}
        \caption{WeedMap.}
        \label{fig:acc_vs_gsd}
      \end{subfigure}
     \hfill
      \begin{subfigure}[b]{\linewidth}
        \centering
         \includegraphics[width=0.9\linewidth]{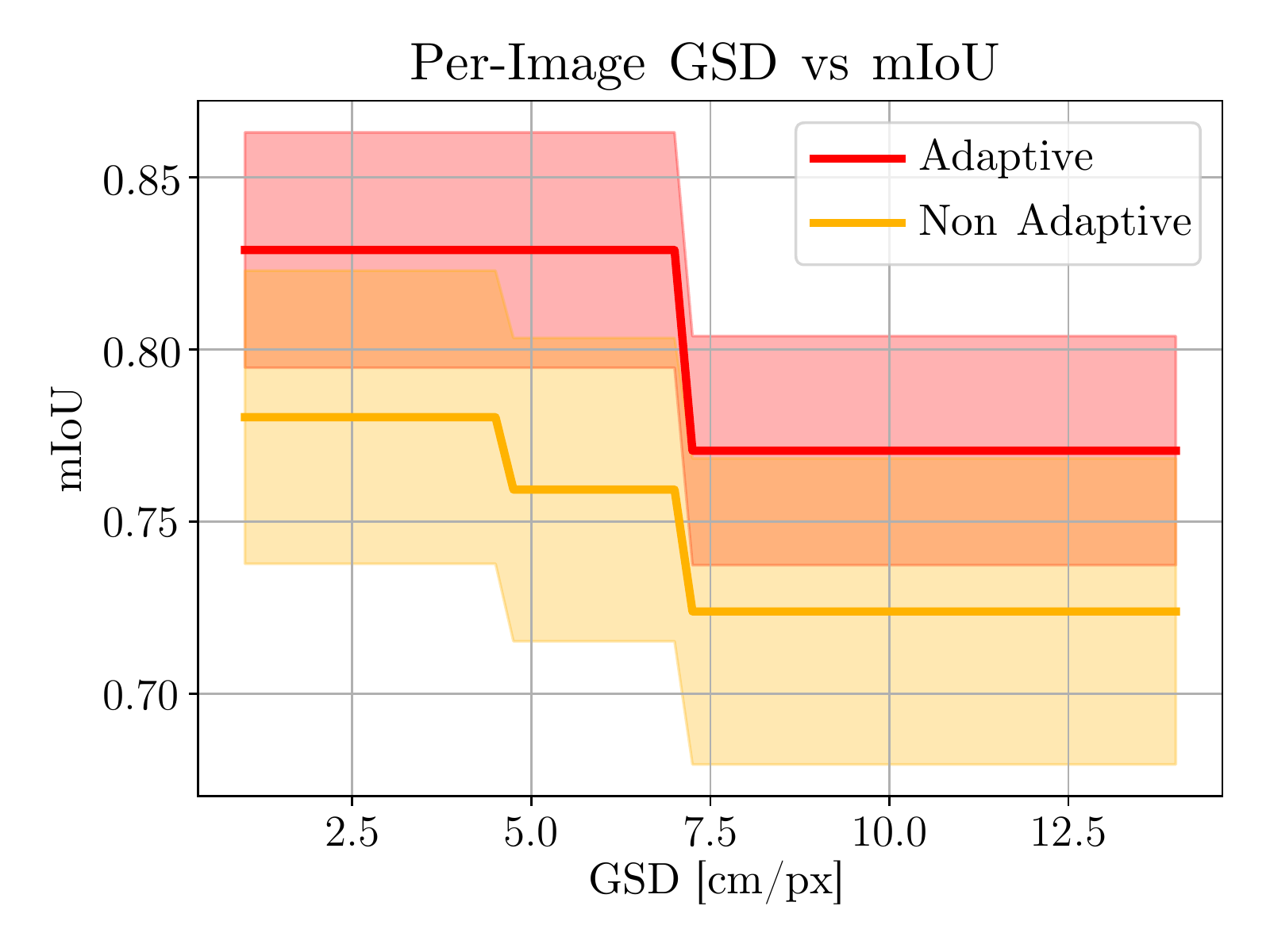}
        \caption{RIT-18.}
        \label{fig:acc_vs_gsd_rit}
      \end{subfigure}
      \caption{Means and standard deviations of the per-image statistics for semantic segmentation. Our adaptive strategy leads to better performance when scouting the field at low altitudes (high GSDs).}
      \label{fig:iouvsh}
\end{figure}

The first experiment is designed to show that our proposed strategy obtains higher accuracy when compared against the baseline methods while keeping low execution time. We show such results in \figref{fig:results} for WeedMap and \figref{fig:results_rit} for RIT-18. For each strategy, we compute the mIoU (over the entire field) and the execution time needed by the UAV to complete its path. The adaptive strategy crosses the line defined by the lawnmower strategies at different altitudes, meaning that it can achieve better segmentation accuracy while keeping a lower execution time. The non-adaptive strategy instead lies under the curve, failing to overtake the lawnmower strategy. We plot exemplary paths results from the different strategies in \figref{fig:waypoints}; in the top row on the left, we show the lawnmower strategy with altitudes corresponding to GSDs of $1.0~\text{cm}/\text{px}$ and $3.0~\text{cm}/\text{px}$, in the bottom row the lawnmower strategy corrispond to GSDs of $4.7~\text{cm}/\text{px}$ and $14.1~\text{cm}/\text{px}$. In both cases, the middle and right plots show the paths resulting from non-adaptive and adaptive strategy, respectively.

For the experiment on WeedMap, our set of targeted labels of interest comprises the two vegetation classes, $\mathcal{L}~=~\{ \textit{crop}, \textit{weed}\}$. This is representative of a precision agriculture task where mission objective is to closely inspect plants. Using this subset of labels, the improvement of the mIoU is mainly given by the \textit{crop} class while the accuracy values for the remaining classes remain stable.

For the experiment on RIT-18, we target one class at a time and plot the average results over each run, resulting in a total of 10 runs considering the 5 classes each with the testing and validation sets reversed. %
Results from these experiments show that our approach yields better segmentation performance and lower execution time when the target class is not dominating the scene. Intuitively, the more spatially localized the class is, the more closely it can be inspected in a targeted way and hence the greater the benefit of using our adaptive multi-resolution strategy to reduce altitude only in this area.
This is the case, for example, when targeting classes such as \textit{ashpalt} or \textit{water} in RIT-18.
As qualitative examples, \figref{fig:qualitative} compares the semantic masks obtained using the adaptive approach and a lawnmower strategy. The circled details illustrate situations where our proposed adaptive method produces visually more correct segmentations without loss of detail by allowing the UAV to reobserve a target area at higher resolutions.

\subsection{Per-Image Segmentation Accuracy vs. Altitude}
\label{sec:semseg_acc_alt}
The second experiment shows the ability of our approach to achieve targeted semantic segmentation when compared to the non-adaptive strategy. At this stage, we compute mIoU for each image that contributes to the final segmentation of the whole field. This gives us a way of evaluating the efficiency of our adaptation strategy. We then visualize the means and standard deviations. As can be seen in \figref{fig:iouvsh}, our adaptive strategy provides higher per-image accuracies when the UAV is scouting the field at low altitudes. This entails that, with our strategy, the limited flight time resources are spent in a more efficient manner in terms of monitoring performance.

\section{Conclusion}
\label{sec:conclusion}

This paper presents a new approach for adaptive path planning for UAVs in general multi-resolution semantic segmentation applications. A key contribution of this paper is a new adaptive planning algorithm that directly tackles the altitude dependency of the deep learning semantic segmentation model using UAV-based imagery. Our strategy exploits the prior knowledge of a field and the new incoming segmentations to enable adaptive decision-making for mapping targeted areas of interest at higher image resolutions. Experimental results using real-world data from different domains validate that our strategy leads to high segmentation accuracy while minimizing flight time needed to cover the field. Our approach opens a direction for efficient UAV mapping, especially in applications such as precision agriculture, where certain areas of a field need to be closely inspected.

\bibliographystyle{cas-model2-names}
\bibliography{glorified, new}

\end{document}